\documentclass[letterpaper, 10 pt, conference]{ieeeconf}  

\IEEEoverridecommandlockouts                              

\usepackage{graphicx}
\usepackage{cite}
\usepackage{amsmath,amssymb,amsfonts}
\usepackage{algorithm}
\usepackage{algorithmic}
\usepackage{comment}
\usepackage{caption}
\usepackage{subcaption}
\usepackage{multirow}
\usepackage{float}    
\usepackage{placeins} 
\usepackage{hyperref}
\usepackage{booktabs}
\usepackage{svg}
\usepackage{eso-pic}
\newcommand{\mycopyrighttext}{%
  \footnotesize
  \noindent
  \textcopyright~2025 IEEE. Personal use of this material is permitted. Permission from IEEE must be obtained for all other uses, in any current or future media, including reprinting/republishing this material for advertising or promotional purposes, creating new collective works, for resale or redistribution to servers or lists, or reuse of any copyrighted component of this work in other works.\\
  IEEE 36th Intelligent Vehicles Symposium (IV 2025) - 22-25 June, 2025.
}

\AtBeginDocument{%
  \AddToShipoutPicture*{%
    \AtPageUpperLeft{%
      \put(55, -40){
        \parbox{\dimexpr\textwidth-4pt\relax}{\raggedright \mycopyrighttext}
      }%
    }
  }
}

\title{\LARGE \bf Boundary-Guided Trajectory Prediction for Road Aware and Physically Feasible Autonomous Driving}
\author{Ahmed Abouelazm$^{1}$, Mianzhi Liu$^{2}$, Christian Hubschneider$^{1,2}$,\\ Yin Wu$^{2,3}$, Daniel Slieter$^{3}$, and J. Marius Zöllner$^{1,2}$
\thanks{$^{1}$Authors are with the FZI Research Center for Information Technology, Germany
        {\tt\small abouelazm@fzi.de}}%
\thanks{$^{2}$Authors are with the Karlsruhe Institute of Technology, Germany}%
\thanks{$^{3}$Authors are with CARIAD SE, Germany}%
}

\begin{document}
\maketitle
\thispagestyle{empty}
\pagestyle{empty}

\begin{abstract}
    Accurate prediction of surrounding road users' trajectories is essential for safe and efficient autonomous driving. While deep learning models have improved performance, challenges remain in preventing off-road predictions and ensuring kinematic feasibility. Existing methods incorporate road-awareness modules and enforce kinematic constraints but lack plausibility guarantees and often introduce trade-offs in complexity and flexibility. This paper proposes a novel framework that formulates trajectory prediction as a constrained regression guided by permissible driving directions and their boundaries. Using the agent’s current state and an HD map, our approach defines the valid boundaries and ensures on-road predictions by training the network to learn superimposed paths between left and right boundary polylines. To guarantee feasibility, the model predicts acceleration profiles that determine the vehicle’s travel distance along these paths while adhering to kinematic constraints. We evaluate our approach on the Argoverse-2 dataset against the HPTR baseline. Our approach shows a slight decrease in benchmark metrics compared to HPTR but notably improves final displacement error and eliminates infeasible trajectories. Moreover, the proposed approach has a superior generalization to less prevalent maneuvers and unseen out-of-distribution scenarios, reducing the off-road rate under adversarial attacks from 66\% to just 1\%. These results highlight the effectiveness of our approach in generating feasible and robust predictions.
    
    \begin{keywords}
    Trajectory Prediction, Motion Planning
    \end{keywords} 
\end{abstract}
\section{Introduction}
\label{sec:Introduction}
Trajectory prediction is a vital component of decision-making in autonomous driving, bridging perception and planning by forecasting the future states of surrounding road users~\cite{gupta2021deep,Caesar2021nuPlanAC}. These predictions enable planning modules to navigate safely and avoid collisions in complex environments~\cite{sun2020scalability}. Traditional trajectory prediction relies on rule-based systems with predefined heuristics from traffic rules and expert knowledge~\cite{rudenko2020}. While these systems are interpretable and straightforward, their rigidity makes them poorly suited for the variability of real-world behaviors.

Deep learning has transformed trajectory prediction by addressing these limitations. By learning complex patterns and contextual information directly from data, these models capture nuanced road user behaviors and adapt to dynamic scenarios~\cite{cui2019multimodal}. Multi-modal deep learning approaches further enhance performance by generating multiple possible trajectories along with their probabilities, enabling robust predictions~\cite{gao2020vectornet}. By leveraging high-definition (HD) maps with detailed road geometry and partial traffic rules information~\cite{liang2020learning,grimm2023holistic}, these models achieve state-of-the-art accuracy.
\begin{figure}[t]
    \centering
    \includegraphics[trim={6cm 5cm 6cm 5.8cm},clip,width=\linewidth]{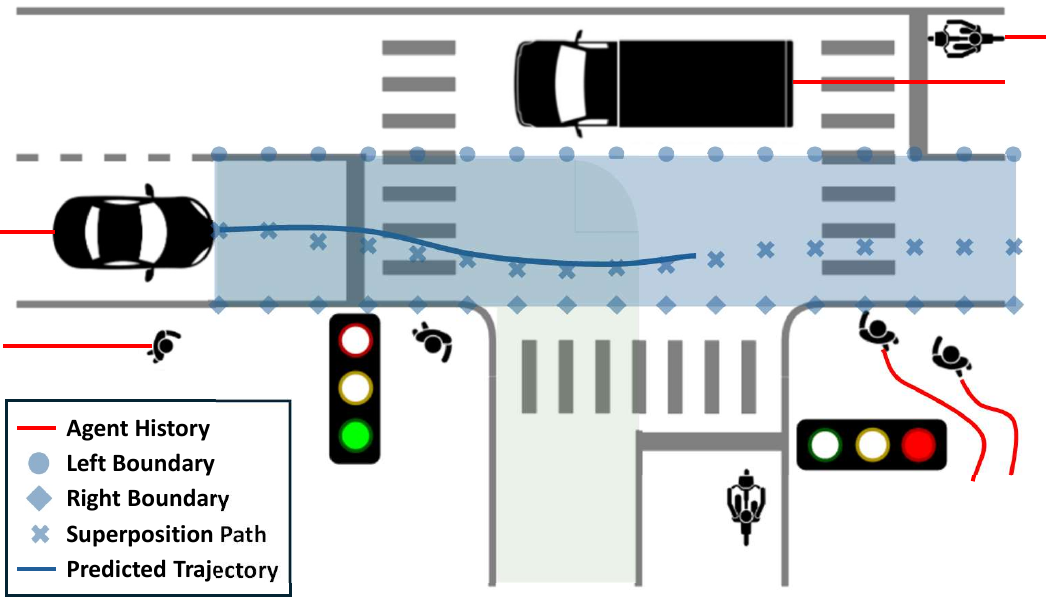}
    \caption{Extracted boundaries for forward (blue) and right (green) driving directions. Reachable lanes are identified based on the agent's state, and equidistant points are sampled along the left and right boundaries. The network learns superposition weights between each driving direction's left and right boundaries to generate a superposition path. Furthermore, the network estimates an acceleration profile to transform the superposition path into a trajectory. The right driving direction predictions are omitted for simplicity.}
    \label{fig:superposition_concept}
    \vspace{-0.35cm}
\end{figure}

\vspace{-0.5em}
\textbf{Research Gap.} Despite advances in deep learning for trajectory prediction, challenges remain in generalization to unseen scenarios and adherence to physical motion constraints. Bahari et al.~\cite{bahari2022vehicle} investigated the challenge of generalizing to unseen road topologies by introducing a scene-perturbation technique called \textit{scene attack}. 
Their findings revealed that state-of-the-art models, such as LaneGCN~\cite{liang2020learning}
, exhibited unexpected fragility to minor changes in road topologies, with over 60\% of the predicted trajectories going off-road. 

Goal-conditioned approaches~\cite{zhao2021tnt} attempt to mitigate the aforementioned challenge by incorporating goal points to enhance road awareness; however, these approaches condition solely on the final point of the predicted trajectory, overlooking the dynamics and continuity of the remaining trajectory. On the other hand, set-based methods~\cite{phan2020covernet} proposed framing trajectory prediction as a classification task over a set of possible trajectories derived from ground truth data. While these methods improve road awareness, they have limited resolution and flexibility. 

Additionally, Girase et al.~\cite{girase2021physically} observed that many deep learning models fail to adhere to the road users' motion constraints, resulting in physically infeasible predictions that overlook steering and acceleration limits. Thus, improving road awareness and generalization to unseen topologies while maintaining physical feasibility remains an open challenge. 

\textbf{Contribution. }This work introduces a novel trajectory prediction framework that integrates the boundaries of permissible driving directions to constraint trajectory regression, ensuring feasibility and better generalization to unseen road topologies. The key contributions are:
\begin{itemize} 
    \item \textbf{Algorithm Development} for extracting permissible driving directions and their boundaries from HD maps.
    \item \textbf{Trajectory Prediction Network Design} that leverages extracted boundaries to predict superimposed paths via the superposition of left and right boundary points.
    \item \textbf{Trajectory Refinement} through a Pure Pursuit layer that transforms superimposed paths and acceleration profiles into physically feasible, on-road trajectories.
\end{itemize}
\begin{figure*}[t]
    \centering
    \includegraphics[trim={0.5cm 5.1cm 0.5cm 5.1cm},clip,width=\textwidth]{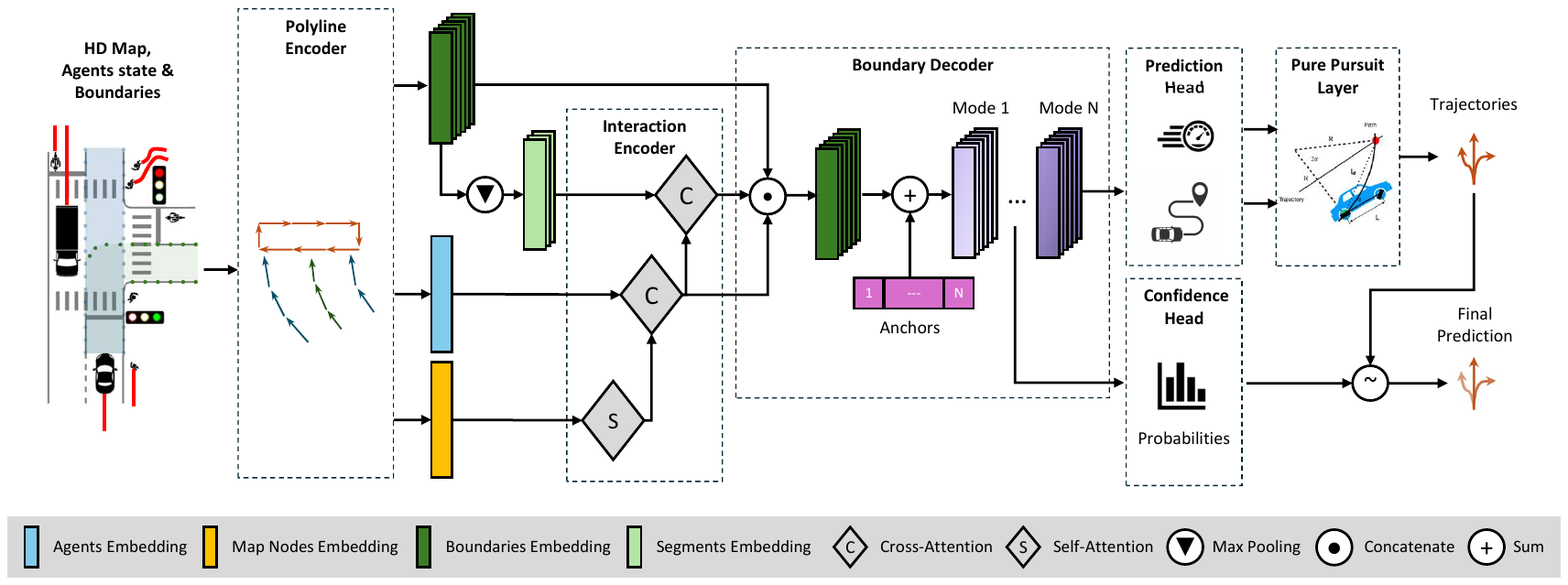}
    \caption{We extend the scene representation by incorporating the generated boundary set. Each boundary consists of a pair of the leftmost and rightmost boundary lanes for a permissible driving direction, sampled at equal distance and represented as a polyline. Scene elements are encoded independently using a polyline encoder, and interactions between them are captured by a transformer-based encoder. The focal agent embedding is then concatenated with each boundary embedding and passed through a boundary decoder to generate mode-specific embeddings. A prediction head subsequently processes these embeddings to compute superposition weights and acceleration profiles for each embedding. Finally, the Pure Pursuit algorithm converts these outputs into feasible trajectories.}
    \label{fig:network}
    \vspace{-0.35cm}
\end{figure*}
\section{related work}
This section reviews the trajectory prediction literature, emphasizing models incorporating kinematic feasibility and enhanced road awareness. 
\subsection{Physical Feasibility}
Ensuring physical feasibility is critical for safe and reliable trajectory prediction~\cite{janjovs2021self}. Infeasible predictions, including unrealistic accelerations or curvatures, stem from the network's lack of motion constraints awareness and noise in the training data~\cite{cui2020deep}. To mitigate these issues, various approaches incorporate feasibility constraints at different stages of the prediction pipeline. 
\textbf{Post-processing} approaches, such as SafetyNet~\cite{vitelli2022safetynet}, assess trajectory feasibility and replace infeasible ones with rule-based fallback trajectories. However, these fallback trajectories often underperform compared to deep learning models~\cite{golchoubian2023pedestrian} and require expert knowledge.

A more \textbf{integrated} approach incorporates feasibility into network learning by predicting control actions such as steering and acceleration and utilizes a kinematic model specific to the agent type to generate feasible trajectories. DKM~\cite{cui2020deep} employs a bicycle model for vehicle trajectory prediction, estimating parameters such as the wheelbase from motion history, though this estimation can introduce noise and degrade prediction quality. Alternatively, Trajectron++~\cite{salzmann2020trajectron++} uses a unicycle model that incorporates non-holonomic constraints without requiring additional parameter estimation.
\subsection{Road Awareness}
Road awareness is essential for trajectory prediction, simplifying scenario interpretation, and ensuring alignment with road topology and traffic rules. A lack of awareness can lead to off-road predictions, as demonstrated in scene attack~\cite{bahari2022vehicle}. 
A promising approach to improving road awareness is \textbf{goal-conditioned prediction}. For instance, MTR~\cite{shi2022motion} and MTR++\cite{shi2024mtr++} iteratively refine their predictions based on rule-based goal candidates. However, their approach lacks an explicit regulation mechanism or loss function to ensure that goal candidates influence the predicted trajectories. TNT~\cite{zhao2021tnt} incorporates goal candidates more effectively with a loss function that ensures the final trajectory point matches a goal candidate. 
However, TNT constrains only the final trajectory point, neglecting the rest of the trajectory and failing to enforce road awareness along intermediate points.

On the other hand, \textbf{set-based prediction} reformulates trajectory prediction as a classification problem, selecting the most plausible trajectory from a predefined set. CoverNet~\cite{phan2020covernet} applied a coverage algorithm~\cite{branicky2008path} to extract diverse trajectory subsets from training data. However, the network performed classification without explicit knowledge of the trajectory subset, as its high dimensionality prevented inclusion in the network input. KI-PMF~\cite{vivekanandan2024ki} addressed this limitation by introducing a refinement layer to remove physically infeasible and off-road trajectories, along with an efficient transformer-based encoder to integrate the trajectory set into the network input. Yet, these methods can only select trajectories from the training data, limiting their generalization to unseen road topologies.

PRIME~\cite{song2022learning} addressed this by extracting a scene-specific trajectory set in \textbf{Frenet space}, followed by a similar approach in~\cite{hallgarten2024stay}. However, these approaches represent each lane in a distinct Frenet space, requiring separate forward passes for each reachable lane. This leads to computational overhead and challenges in modeling complex maneuvers like lane changes. Overall, set-based methods can be hindered by the limited resolution of generated sets and the complexity of solving classification problems over thousands of trajectories, as opposed to regression over fewer modes.

These limitations inspired an algorithm to extract the boundaries of permissible driving directions per scene, overcoming the single-lane constraints of Frenet-based approaches. Additionally, left and right boundary points constrain trajectories on-road, reformulating the problem as a constrained regression task with infinite resolution, overcoming set-based limitations.
\section{Methodology}

In this work, we propose a network architecture, illustrated in Fig.~\ref{fig:network}, that enhances road awareness while ensuring physical feasibility. The network leverages boundaries of permissible driving directions determined by the focal agent's current state within an HD map. These boundaries, represented as the outermost left and right edges of a driving direction~\cite{karle2023mixnet}, are encoded as polylines and enriched with attention mechanisms to incorporate information about surrounding agents and map nodes. Using the learned boundary embeddings, a prediction head generates multi-modal superposition weights and acceleration profiles between the left and right boundaries. Finally, a pure pursuit kinematic layer converts these weights and acceleration profiles into a trajectory for each mode. The following sections describe the algorithm for extracting driving boundaries and the network architecture designed to integrate boundary information.

\subsection{Boundary Set Generation}
\begin{figure*}[h]
    \centering
    \begin{subfigure}[t]{0.325\linewidth}
    \centering
    \includegraphics[width=\linewidth]{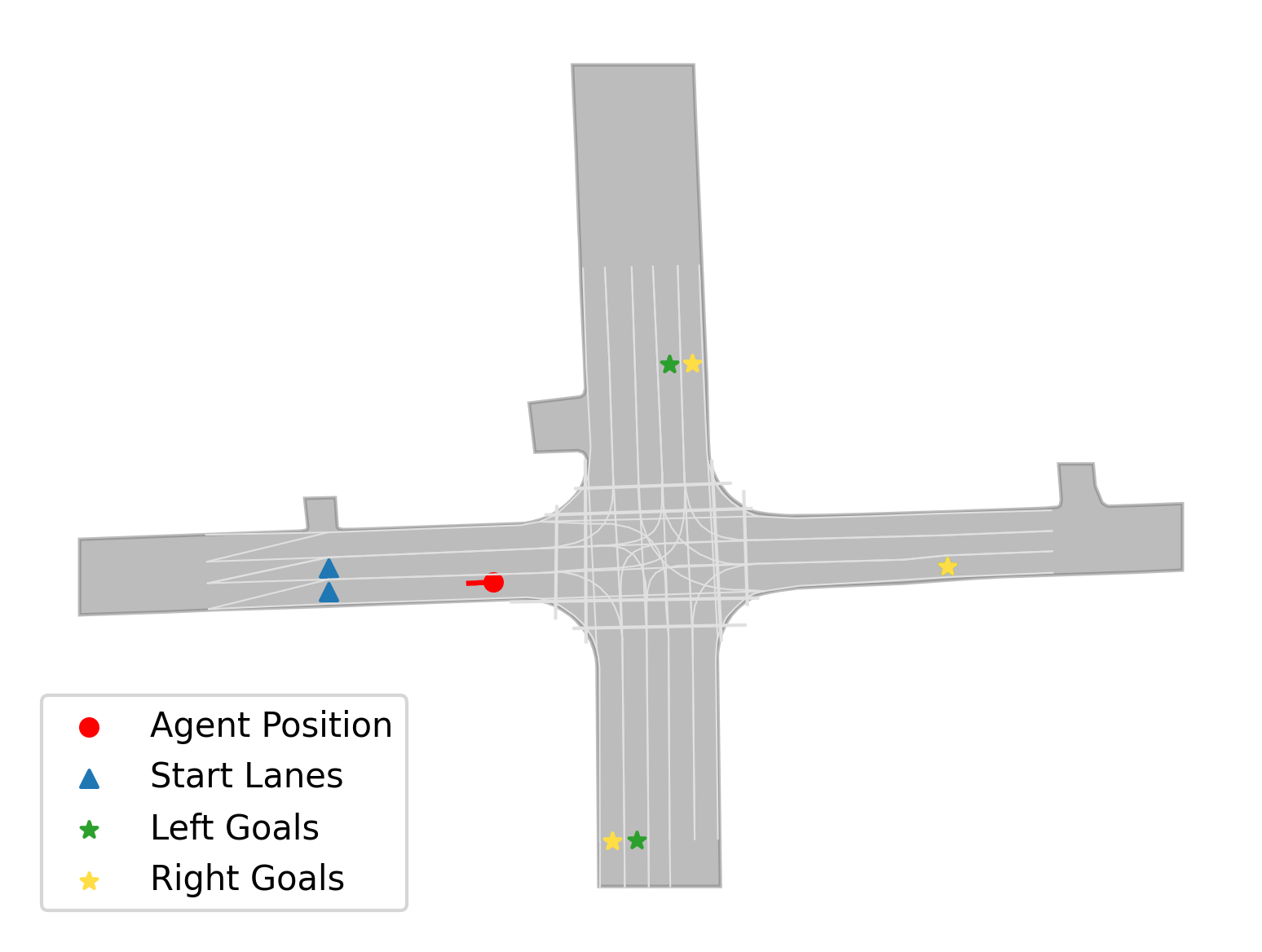}
    \caption{Identify potential start and goal lanes based on the agent state.}
    \label{fig:boundary_set_start}
    \end{subfigure}
    \hfill
    \begin{subfigure}[t]{0.325\linewidth} 
    \centering
    \includegraphics[width=\linewidth]{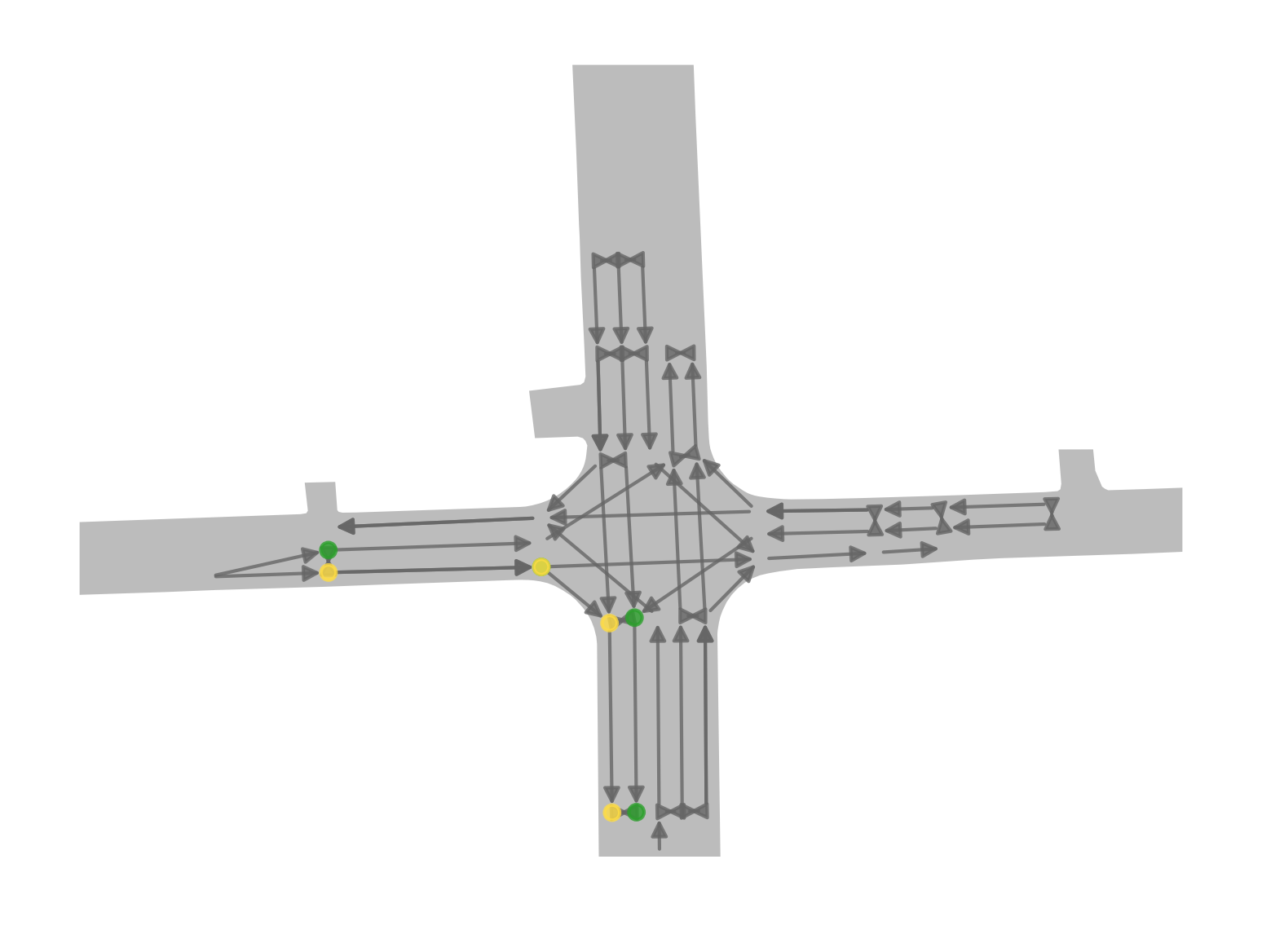}
    \caption{Graph representation of HD map and extracted boundary nodes for right driving direction. Other driving directions are omitted for simplicity.}
    \label{fig:boundary_set_search}
    \end{subfigure}
    \hfill
    \begin{subfigure}[t]{0.325\linewidth}
    \centering
    \includegraphics[width=\linewidth]{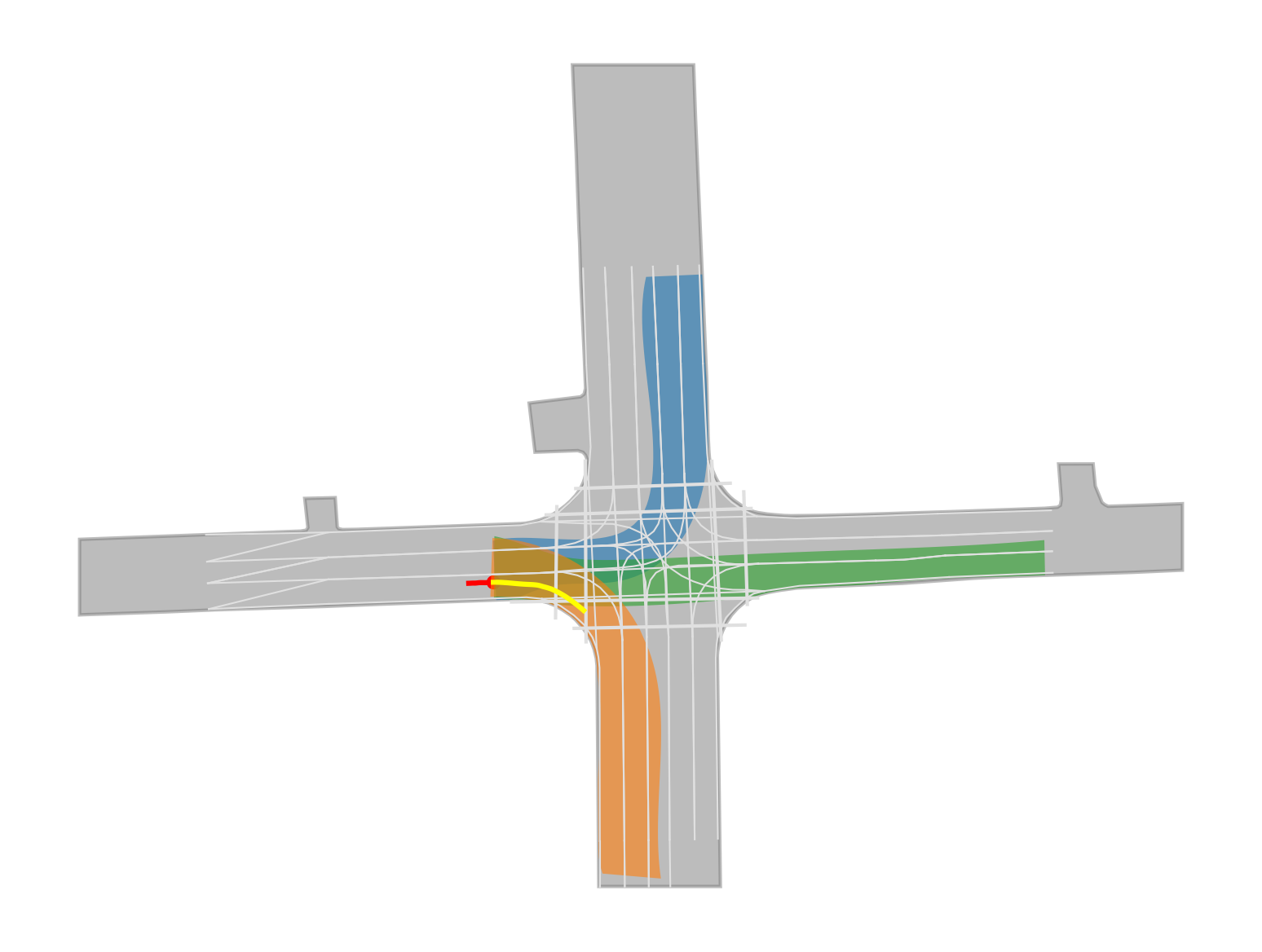}
    \caption{The generated boundary set with all permissible driving directions.}
    \label{fig:boundary_set_final}
    \end{subfigure}
    \caption{Our boundary generation algorithm represents the HD map as a graph and identifies the nearest start lanes based on the agent's current state. It then performs reachability analysis to determine goal lanes, including the leftmost and rightmost goal lanes in each driving direction. Next, a modified depth-first search algorithm is used to extract the leftmost and rightmost lanes traversed from the start to the goal. Finally, the boundary lanes are smoothed and uniformly sampled at equidistant intervals.}
    \label{fig:boundary_set_example}
     \vspace{-0.35cm}
\end{figure*}

To enhance road awareness and ensure adherence to road topology, we introduced the concept of a boundary set comprising up to $N_b$ boundaries. Each boundary is defined by left and right polylines, representing permissible driving areas for all possible directions for the focal agent. These boundaries are generated by the following steps:

\begin{enumerate}
    \item \textbf{Identify potential start lanes} based on the focal agent's position and heading, as illustrated in Fig.~\ref{fig:boundary_set_start}.
    
    \item \textbf{HD map representation} as a directed graph where lanes are nodes connected by directional edges (e.g., successor, predecessor, left, right) similar to~\cite{liang2020learning}.
    
    \item \textbf{Reachability analysis} to identify goal lanes reachable from the start lanes. Cluster these goal lanes based on their directional edges, then extract the leftmost and rightmost goal lanes in each cluster.

    \item \textbf{Boundary construction} constructs the left boundary using a modified depth-first search, which prioritizes left edges to find a path from the start lane to the leftmost goal lane. 
    The same process is applied to the right boundary, with the right edges prioritized, as shown in Fig.~\ref{fig:boundary_set_search}.

    \item \textbf{Boundary sampling and smoothing} involves sampling the extracted boundaries at equal distances and applying cubic spline smoothing. The final smoothed boundaries are shown in Fig.~\ref{fig:boundary_set_final}.

    \item \textbf{Boundary selection} reduces the number of boundaries if they exceed the desired $N_b$. Non-maximum suppression (NMS) is applied to retain the $N_b$ most relevant boundaries based on the intersection area between different boundaries.
\end{enumerate}

The boundary set has several advantages compared to previous approaches. It is computed per scene using efficient graph traversal algorithms rather than relying on bagging the entire training dataset as in set-based methods. Boundaries provide a more informative constraint than goal points, with superposition between left and right boundary points ensuring the entire trajectory stays within the defined road boundaries. Moreover, grouping lanes with a common driving direction is more general and flexible than single-lane Frenet-based approaches, allowing the network to learn complex maneuvers like lane changes freely.

\subsection{Network Architecture}
We chose HPTR~\cite{zhang2024real} as the backbone of our model for its state-of-the-art and flexible transformer-based architecture. Additionally, HPTR incorporates an efficient attention mechanism called KNARPE. For each token, such as map polylines and agents, KNARPE aggregates the local context by considering the K-nearest neighbors based on $L_2$ distance. 
 
The network input is extended with a boundary set containing up to $N_b$ boundaries, each consisting of a pair of left-right polylines. Each polyline has $N_p$ points sampled at equal distances. We denote the boundary set as $BS \in \mathbb{R}^{N_b \times 2 \times N_p \times 2}$. Compared to set-based approaches~\cite{phan2020covernet} with a trajectory set $TS \in \mathbb{R}^{N_T \times N_p \times 2}$, our method is more computationally efficient since $2 \times N_b \ll N_T$. For example, we use up to $N_b = 6$ boundaries per scene, whereas CoverNet handles $N_T$ in the thousands. This reduced dimensionality enables efficient encoding of the boundary set into the network input.

\textbf{Pre-processing Step.} All network inputs, including agents' history, map lanes, and boundary set, are represented as pairwise relative polylines~\cite{jia2023hdgt}. Each polyline consists of consecutive vectors defined by their global pose (position and heading) and local attributes. 
This viewpoint-invariant representation enhances network robustness and reduces the need for data augmentation during training~\cite{zhang2024real}. Each boundary polyline (left or right) is divided into segments with equal points $N_{seg}$. The global pose encodes the 2D position and heading of the first point in each segment, while local attributes capture each point's displacement and heading change relative to the segment's first point. Segmenting boundaries allows the network to learn fine-grained features, avoiding the limitations of encoding long boundaries with a single global pose.

\textbf{Polyline Encoder.} After pre-processing, the model inputs are passed to a polyline encoder, consisting of separate PointNet~\cite{qi2017pointnet} layers for independent processing of different polyline types. 
PointNet extracts features from local attributes of each point in the polyline using a multi-layer perceptron (MLP), ensuring that these local attributes are captured independently of point order. It then aggregates the extracted features with a linear layer and max pooling, summarizing the key global features across the polyline. The pooled global features are concatenated with the original point-wise features, and this process—reducing dimensionality from $F_e$ to $F_e / 2$ and restoring it to $F_e$—is repeated $N_e$ times to refine feature representations progressively.

To improve efficiency, we apply max pooling after the polyline encoder to consolidate each map and agent polyline into a single token. For the boundary segments, we maintain two versions: the \textbf{boundary embedding} and the \textbf{boundary segment embedding}. The boundary embedding preserves the number of points $N_p$ to retain point-wise fea needed for superposition and acceleration estimation in the prediction head. In contrast, the boundary segment embedding pools each segment into a token, which is passed to the interaction encoder to learn interactions with map and agent tokens.
\textbf{Interaction Encoder.} The interaction encoder is a transformer-based architecture designed to encode interactions both within and across token classes, utilizing the KNARPE mechansim~\cite{zhang2024real} instead of standard attention. 
Map tokens are first enhanced through a self-attention block of KNARPE layers to encode spatial relations within their class. Subsequently, agent tokens are enhanced through a self-attention block to capture inter-agent interactions and further refined by a cross-attention block attending to map tokens. Finally, boundary segment embeddings are enhanced by a self-attention block applied on segment tokens within the same boundary and further refined by a cross-attention block attending to map and agent tokens.

This hierarchical structure emphasizes that while map tokens influence agent behavior, the reverse does not hold. Both map and agent tokens contribute to boundary interpretation—map tokens provide spatial context, while agent tokens capture motion history and intentions relative to boundary points. This approach ensures that boundary segment embeddings incorporate all relevant information, facilitating precise predictions of superposition weights and acceleration profiles within the prediction head.

\subsection{Boundary Decoder}
The boundary decoder utilizes interaction encoder outputs to learn enriched multi-modal representations for each agent’s boundaries, enabling the prediction of multiple trajectories per boundary. Each boundary polyline has two distinct embeddings: the boundary embedding, retaining point-level features derived from the polyline encoder, and the boundary segment embedding, capturing high-level features and contextual interactions with map and agent tokens. To seamlessly incorporate these two sources of embedding, each boundary point embedding is augmented with its corresponding segment embedding, preserving point-specific details while encoding broader contextual information. This enriched embedding is further combined with its relevant agent's embedding, which encodes motion history and interactions with other agents and the map. To reduce dimensionality, an MLP refines the combined embeddings before passing them through a Long Short-Term Memory (LSTM), which captures the sequential nature of boundary polylines.

To incorporate multi-modality into the boundary embedding, the boundary decoder employs $N_{\text{mode}}$ learnable anchor tokens, initialized with high-variance Xavier initialization~\cite{glorot2010understanding}. Learnable anchors enhance mode diversity and mitigate mode collapse~\cite{shi2024mtr++}. The anchors are concatenated with the boundary embedding and passed through an MLP to generate $N_{\text{mode}}$ distinct embeddings.

\subsection{Prediction Head}
The prediction head employs an MLP followed by a softmax to estimate superposition weights for each boundary’s left and right polylines for different modes. This ensures that the weights assigned to a point on the left polyline and its corresponding point on the right sum to one, maintaining the superimposed path on-road. The superimposed path embedding is then passed to an acceleration profile generator, an MLP that predicts acceleration per time step within the prediction horizon. A scaled Tanh function ~\cite{salzmann2020trajectron++} constrains the predicted acceleration within the physically feasible range of $-a_{\text{max}}$ to $a_{\text{max}}$.
\subsection{Pure Pursuit Layer}
We utilize a differential pure pursuit layer~\cite{girase2021physically} to convert a mode-specific superimposed path into an equi-temporal, physically feasible trajectory using the estimated acceleration profile and the vehicle’s initial state. This parameter-free layer updates the vehicle state recurrently and determines the curvature required to reach a goal point on the superimposed path based on the vehicle’s current state and a look-ahead distance $L_d$. The look-ahead distance specifies the forward distance along the path at which the vehicle selects a goal point. The curvature $\kappa_t$ is computed as shown in Eq.~\ref{eq:curvature}, where $x_t^{g}$ is the $x$-coordinate of the goal point in the vehicle’s local coordinate frame. To ensure trajectory feasibility, a maximum curvature limit \(\kappa_{\text{max}}\) is imposed, preventing excessively sharp turns that exceed the vehicle’s physical constraints. Subsequently, the heading update \(\theta_{t+1}\) follows Eq.~\ref{eq:heading_update}, ensuring smooth trajectory updates based on the vehicle’s velocity \(v_t\) and curvature \(\kappa_t\).
\begin{equation}
    \kappa_t = \operatorname{sgn}(x_t^{g}) \cdot \min\left (  \frac{2 \cdot |x_t^{g}|}{L_d^2},\, \kappa_{\max}\right ) \label{eq:curvature}
\end{equation}
\begin{equation}
    \theta_{t+1} = \theta_t + v_t \cdot \kappa_t \cdot \Delta t \label{eq:heading_update}
\end{equation}
Moreover, constraining superimposed paths within road boundaries ensures that the output trajectories remain on the road and comply with legal driving directions. During inference, we apply NMS to reduce redundancy by decreasing the likelihood scores of trajectories whose endpoints are within a distance of $\epsilon$.  This encourages the selection of multiple plausible trajectories with greater diversity.

\subsection{Classification Head}
To estimate the probability of a predicted trajectory, we use an MLP classification head applied to the max-pooled embeddings of its corresponding superimposed path. A softmax layer then normalizes the likelihood scores across all $N_b \times N_{\text{mode}}$ trajectories. We also explored a two-stage classification approach, where a trajectory's likelihood score is computed as the product of its boundary occurrence probability and its probability relative to other modes within the same boundary. While this method can perform well when boundary classification is accurate, errors in boundary classification can strongly bias the model toward incorrect boundaries. 
Due to its greater flexibility, we ultimately chose the first approach.
\begin{figure*}[t]
    \centering
    \begin{subfigure}[t]{0.24\linewidth}
    \centering
    \includegraphics[width=\linewidth]{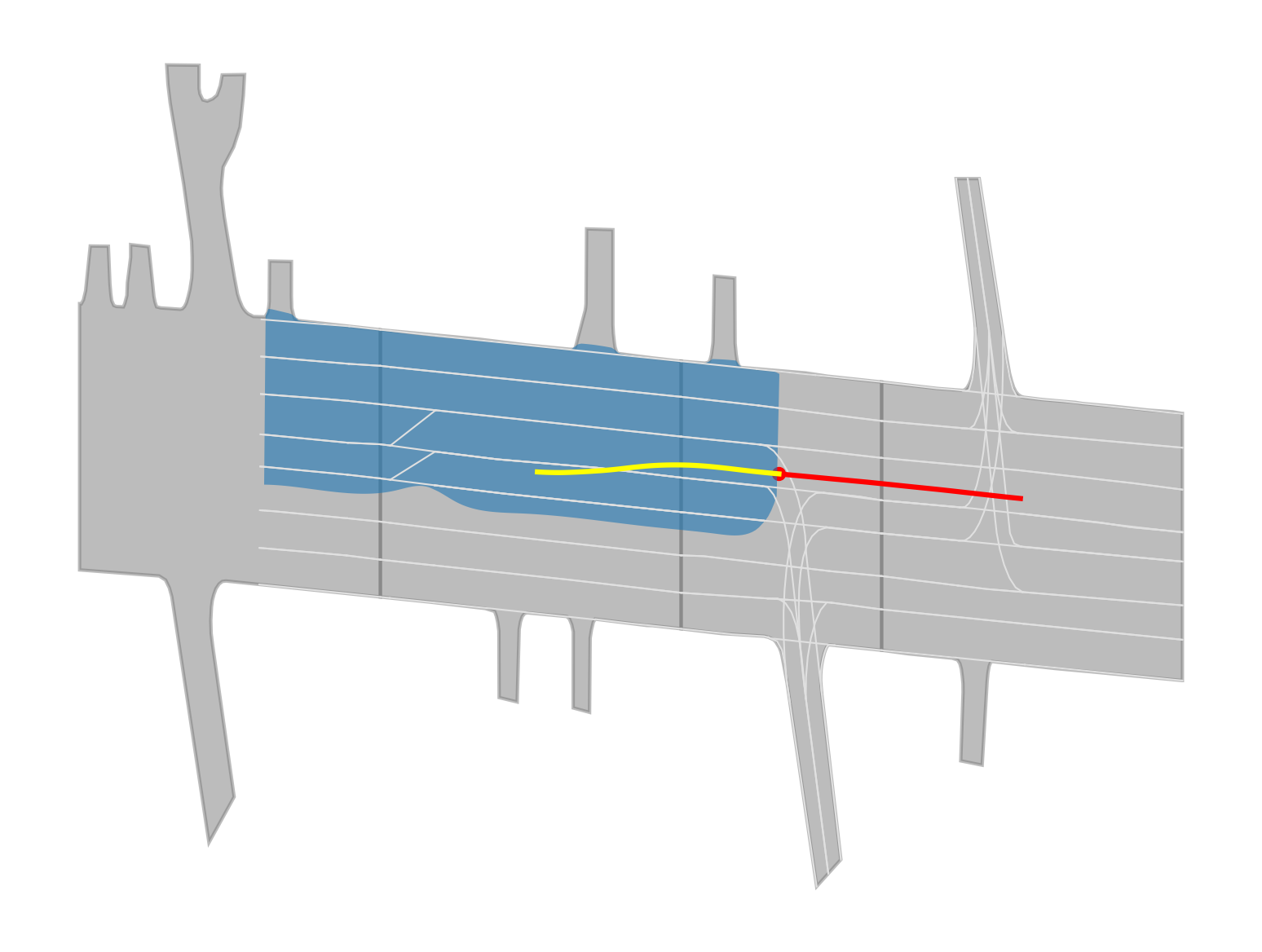}
    \end{subfigure}
    \hfill
    \begin{subfigure}[t]{0.24\linewidth}
    \centering
    \includegraphics[width=\linewidth]{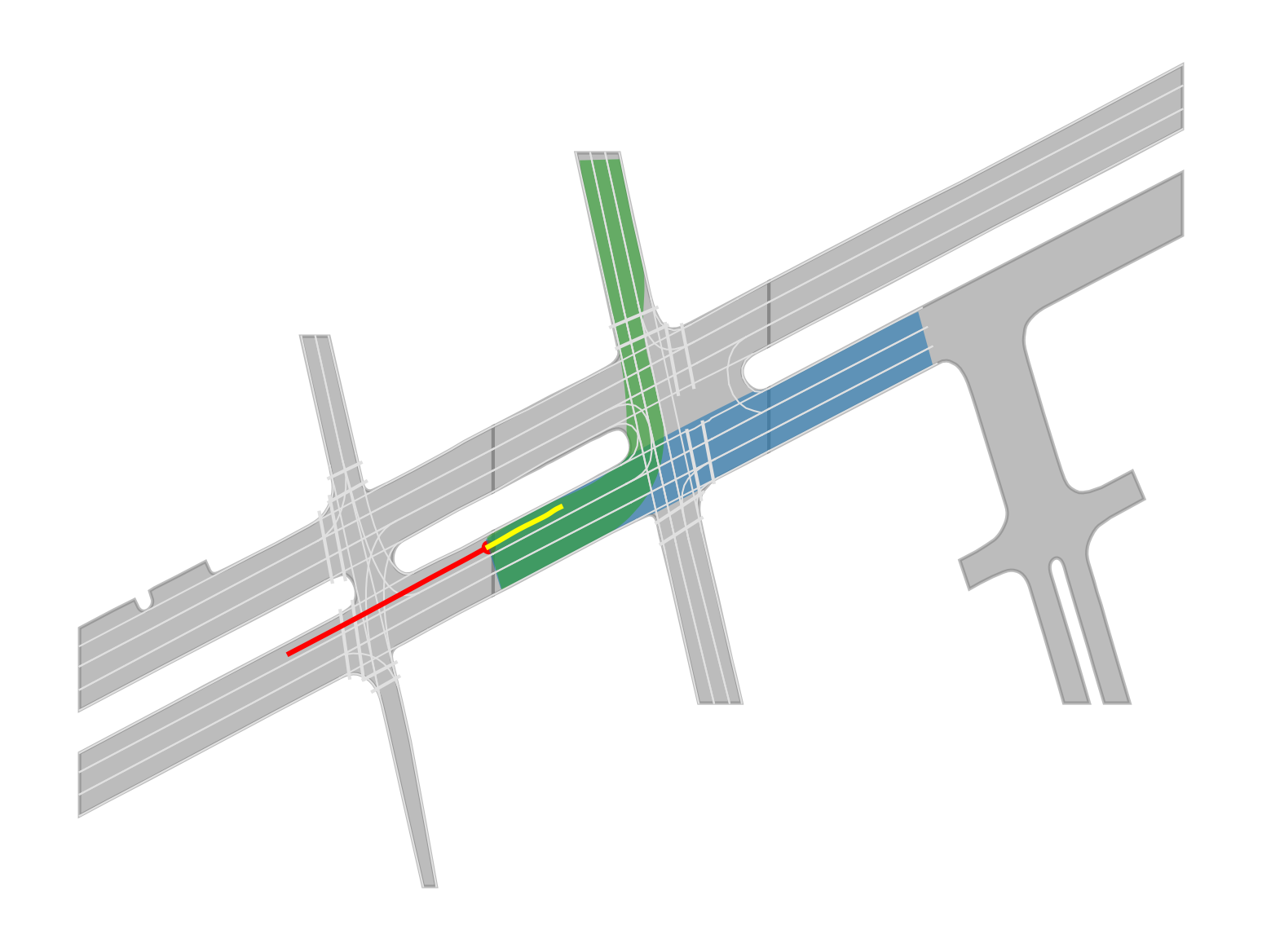}
    \end{subfigure}
    \hfill
    \begin{subfigure}[t]{0.24\linewidth}
    \centering
    \includegraphics[width=\linewidth]{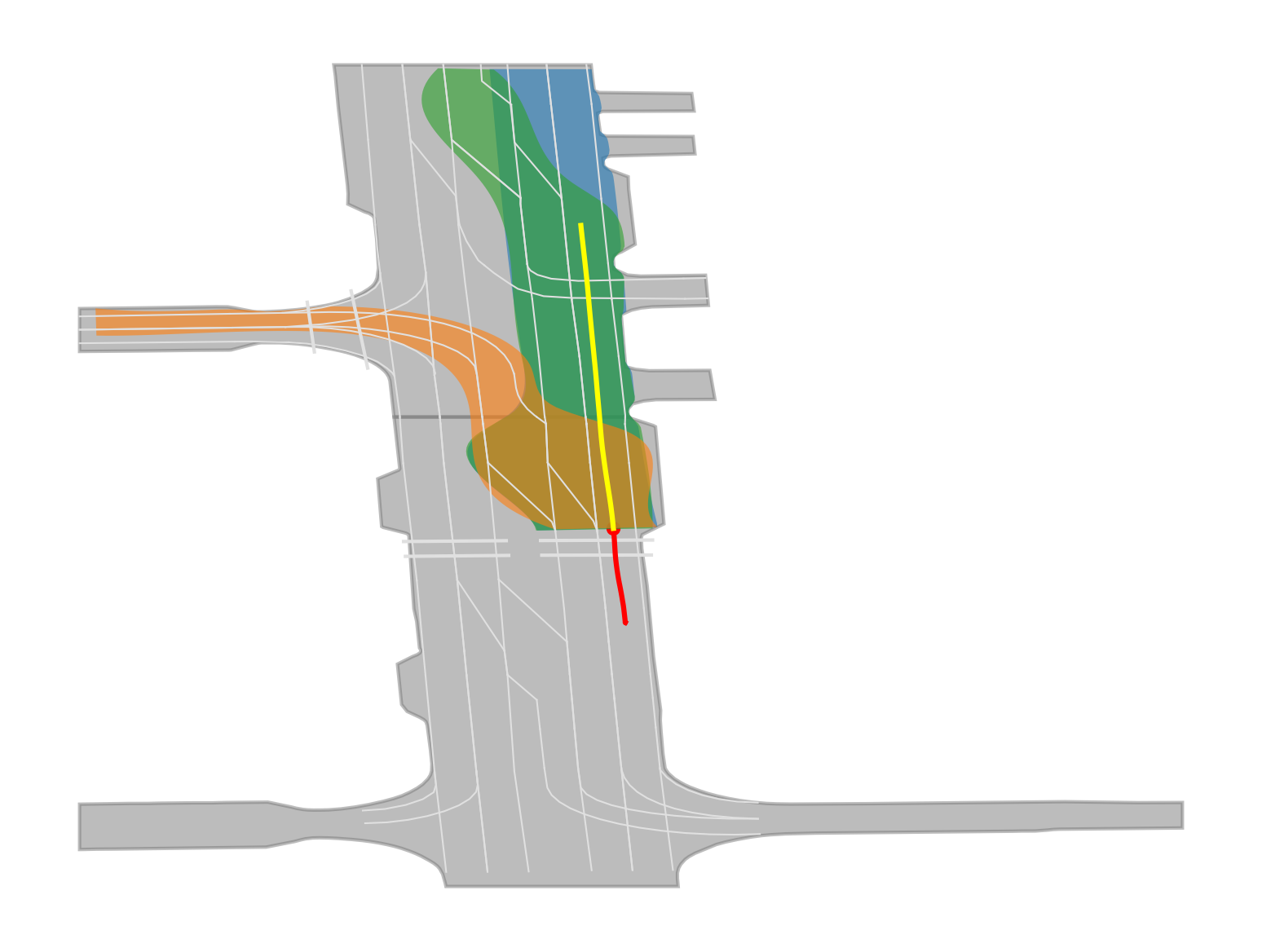}
    \end{subfigure}
    \hfill
    \begin{subfigure}[t]{0.24\linewidth}
    \centering
    \includegraphics[width=\linewidth]{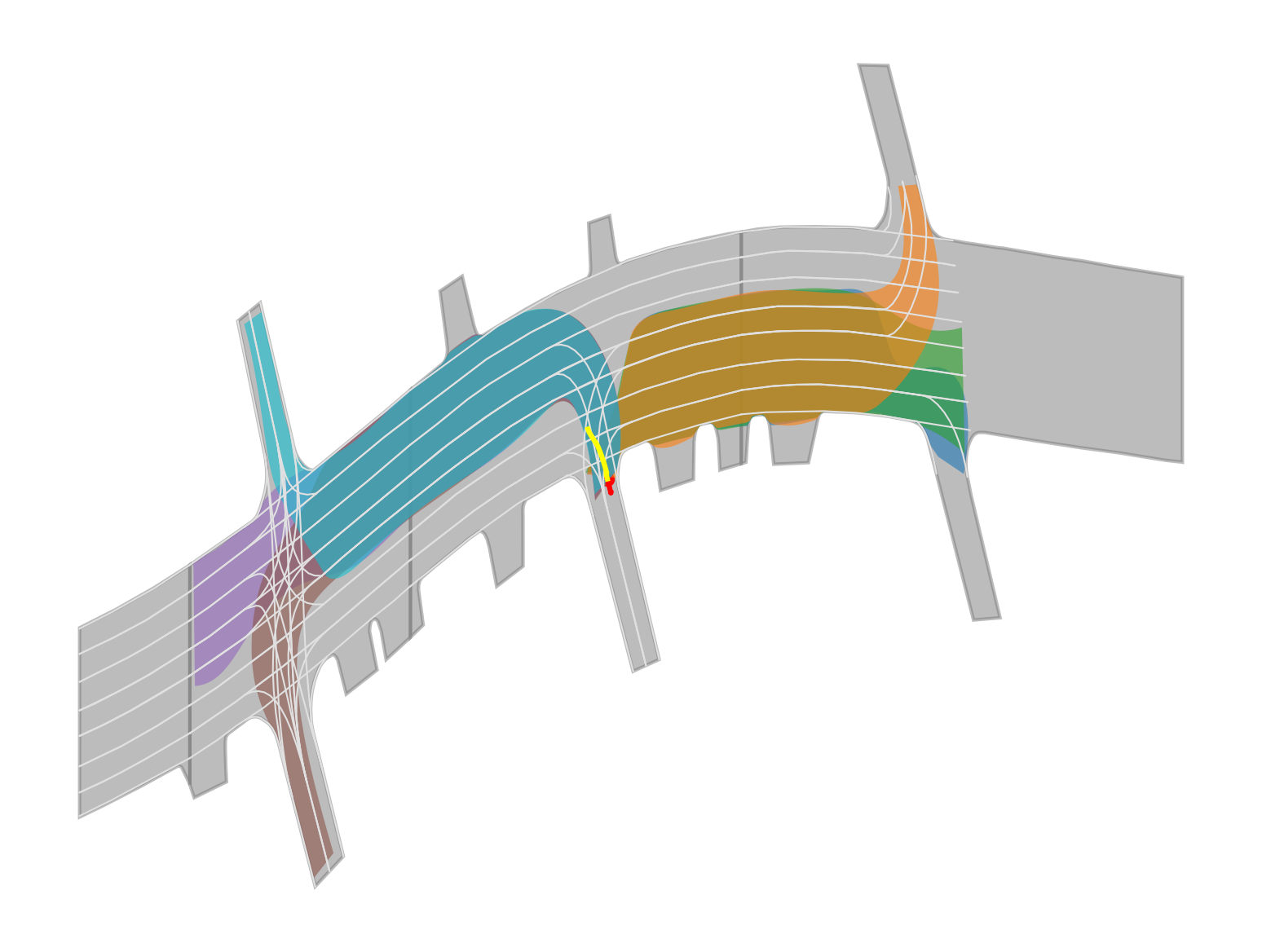}
    \end{subfigure}
    \caption{Examples of boundary sets with varying complexity demonstrate the adaptability of the proposed algorithm. These examples range from simple scenarios with a single boundary to more complex ones with up to six boundaries, showcasing the generation algorithm's ability to handle diverse road layouts and driving directions effectively.}
    \label{fig:boundary_set_success}
    \vspace{-0.35cm}
\end{figure*}

\section{experimental Setup}
This section outlines the experimental setup, covering the dataset, network, and evaluation baselines and metrics.

\textbf{Dataset.} We evaluate our proposed method on the Argoverse-2 (AV2) Forecasting Dataset \cite{wilson2023argoverse}, which contains 250,000 non-overlapping driving scenarios. 
Each scenario spans 11 seconds, recorded at 10 Hz, and provides detailed trajectory data, including position, velocity, and orientation of all observed tracks. The first five seconds serve as the observation horizon, while the subsequent six seconds constitute the prediction horizon, where the model predicts future trajectories of dynamic agents.

\textbf{Network Architecture.} For each scenario, we generate up to \( N_b = 6 \) boundaries to capture diverse plausible driving directions. On average, the AV2 dataset contains 3-4 valid driving directions per scenario. Each boundary polyline, whether left or right, spans up to 150 meters (\( N_p = 150 \)) and is sampled at one-meter distances. During pre-processing, boundary polylines are divided into segments of 5 meters each (\( N_{seg} = 5 \)). We employ a 3-layer PointNet as the polyline encoder, with a hidden dimension \( F_e\) of 256. The interaction encoder applies $K = 36$ for nearest neighbors in self-attention blocks and $K = 144$ for cross-attention blocks. Each attention block includes two sequential KNARPE layers with four heads and a 0.1 dropout rate. 

For each valid boundary, the boundary decoder predicts six distinct modes, reflecting diverse trajectory possibilities. In the prediction head, estimated accelerations are constrained within \([-8, 8]\) m/s\(^2\), as reported in~\cite{girase2021physically}, ensuring feasible motions. A look-ahead distance \( L_d\) of 10 \( \mathrm{m} \) is applied, paired with a curvature constraint \( \kappa_{max}\) of 0.3 \( \mathrm{m}^{-1} \). 
During inference, an NMS distance threshold \(\epsilon = 2 \) ensures balanced trajectory refinement and selection.

\textbf{Loss Function.} The model's loss function consists of two key components: the classification loss and the trajectory regression loss. These components ensure the model predicts both the most probable trajectory and accurately regresses trajectory states—such as position, velocity, and yaw angle—over the prediction horizon. The classification loss is formulated as a cross-entropy loss~\cite{liang2020learning}. For trajectory regression, we adopt Huber loss with a hard assignment strategy, minimizing the distance between the closest predicted trajectory and the ground truth, as measured by average displacement error.

We refrain from incorporating losses that penalize heading and velocity estimation errors~\cite{zhang2024real}, as these states are inherently determined by the pure pursuit layer rather than being independently predicted, as in traditional regression methods. Furthermore, the AV2 dataset exhibits noticeable inaccuracies in velocity and heading data, which could lead the model to optimize for physically infeasible behaviors. We train the model using the AdamW optimizer with an initial learning rate of $10^{-4}$. A learning rate scheduler reduces the learning rate by 50\% every 10 epochs. The model is trained for 30 epochs with a batch size of 8 scenarios.

\textbf{Baselines and metrics.} Given the modular nature of our contribution, ${\text{HPTR}}_{\,\text{bd}}$, our evaluation focuses on its impact within the HPTR backbone, comparing it against different HPTR variants rather than other backbones. The first variant is the original HPTR architecture trained with Huber loss to optimize trajectory predictions. The second, referred to as ${\text{HPTR}}_{\,\text{kin}}$, is a variant of the HPTR that incorporates a unicycle kinematic model~\cite{salzmann2020trajectron++} during training, thereby constraining its predicted trajectories to adhere to physically feasible behavior. To assess trajectory prediction accuracy, we use benchmark metrics such as minimum Average Displacement Error (minADE) and minimum Final Displacement Error (minFDE), both measured in meters, along with Miss Rate (MR). These metrics are widely utilized in literature~\cite{schuetz2023review} and benchmarks such as AV2~\cite{wilson2023argoverse}. Furthermore, we evaluate the physical feasibility of the predictions based on acceleration and curvature limits proposed in~\cite{girase2021physically}. Finally, we assess the model's ability to generalize to unseen road topologies by measuring its off-road performance using the Hard Off-road Rate (HOR), and Soft Off-road Rate (SOR)~\cite{bahari2022vehicle}.
\section{Evaluation}
\begin{table*}
    \centering
    \caption{
    Comparison of plain HPTR, HPTR$_{\,\text{kin}}$, and our model on the AV2 validation split using benchmark metrics. The lower bound reflects the optimal performance of a kinematic model that strictly follows the ground truth under feasibility constraints.}
    \resizebox{0.85\textwidth}{!}{
    \begin{tabular}{l|ccccccc}
    \toprule
        Model & minADE$_6$ & minFDE$_6$ & brier-minADE$_6$ & brier-minFDE$_6$ & minADE$_1$ & minFDE$_1$ & MR$_{2.0}$ \\ \midrule
        Lower Bound & - & - & - & - & 0.21 & 0.59 & 0.02 \\
        HPTR & \textbf{0.88} & \textbf{1.71} & \textbf{1.48} & \textbf{2.32} & \textbf{2.24} & 5.60 & \textbf{0.26} \\
        ${\text{HPTR}}_{\,\text{kin}}$ & 1.10 & 2.01 & 1.71 & 2.62 & 2.50 & 6.12 & 0.31 \\
        ${\text{HPTR}}_{\,\text{bd}}$ (Ours) & 0.95 & 1.82 & 1.60 & 2.47 & 2.27 & \textbf{5.51} & 0.30 \\ \bottomrule
    \end{tabular}}
    \label{tab:compare_one}
    \vspace{-0.2cm}
\end{table*}
In this section, we evaluate the effectiveness of our boundary set generation algorithm in extracting boundaries from an HD map. Furthermore, we assess the impact of incorporating these extracted boundaries into our network architecture, analyzing its performance in terms of accuracy, physical feasibility, and generalization to unseen road topologies.
\subsection{Boundary Set Generation}
The proposed boundary generation algorithm reliably extracts valid boundaries aligned with all permissible driving directions on the AV2 dataset. It effectively handles a diverse range of scenarios, from simple single-boundary cases to more complex scenarios involving six or more boundaries, as illustrated in Fig.~\ref{fig:boundary_set_success}. In rare instances, accounting for 0.28\% of all scenarios in the AV2 dataset, boundary extraction does not succeed due to dataset inconsistencies such as discontinuous drivable areas and missing lane information in HD maps. Since these issues originate from the dataset rather than the algorithm itself, we leave their resolution to future AV2 updates.

\subsection{Prediction Accuracy}
Table~\ref{tab:compare_one} provides a comparison between the proposed model ${\text{HPTR}}_{\,\text{bd}}$ against HPTR and ${\text{HPTR}}_{\,\text{kin}}$ based on benchmark evaluation metrics. We introduce a performance lower bound to provide a more comprehensive assessment of models incorporating kinematic layers. This bound is determined using a kinematic model that tracks ground truth trajectories as closely as possible under physical constraints, representing the best possible performance for models with kinematic layers. The results indicate that integrating feasibility constraints leads to a slight decrease in the performance of ${\text{HPTR}}_{,\text{kin}}$ compared to the original HPTR. This minor reduction can be attributed to the fact that models incorporating kinematic constraints have a lower bound of 0.21 m for minADE and 0.59 m for minFDE. However, the incorporation of boundary-guided prediction mitigates this impact by enhancing road awareness, leading to a noticeable performance recovery. Although the overall performance remains slightly below HPTR, the proposed model shows an improvement in minFDE$_1$, demonstrating better accuracy in predicting goal positions compared to HPTR.

To further evaluate our model's performance, we conducted a maneuver-based analysis of various driving scenarios, including stationary, straight, straight left, straight right, left turn, right turn, and U-turns, as proposed in~\cite{ettinger2021large}. In simpler maneuvers such as stationary, straight, straight left, and straight right, our model performs on par with HPTR. However, in more complex scenarios, such as turns and U-turns, our model outperforms HPTR, as shown in table~\ref{tab:compare_three}, which highlights performance for left U-turn maneuvers. This improvement is largely due to the inclusion of boundary sets, which enable the model to predict trajectories in all permissible driving directions, preventing overfitting to simpler, more frequent cases (e.g., stationary or straight movements) and mitigating the issue of mode collapse. Furthermore, the model's superior performance in U-turn scenarios, which appear only in the validation split and not during training, demonstrates its strong ability to generalize to unseen road topologies compared to HPTR.

\subsection{Physical Feasibility}
The analysis of the infeasibility rate in ground truth data and trajectories predicted by HPTR and our model is detailed in Table~\ref{tab:compare_two}. Results indicate that the AV2 dataset contains a high rate of infeasibility; further discussion of AV2 infeasibility can be found in~\cite{yao2023empirical}. Additionally, HPTR predictions exhibit a significant number of infeasible steps, with 15.6\% of all predicted trajectory steps violating feasibility constraints and 89.1\% of predicted trajectories containing at least one infeasible step. This occurs due to HPTR’s lack of feasibility awareness, leading it to replicate noise from the ground truth. In contrast, integrating the pure pursuit layer in our model completely eliminates infeasibility. While feasibility constraints introduce a trade-off, given the non-zero lower bound and dataset noise, we argue that the slight performance reduction is justified to ensure consistently realistic trajectory predictions.
\begin{table}[]
    \centering
    \caption{
        Infeasibility rate observed in the ground truth data of AV2's validation split, as well as in the trajectories predicted by HPTR and our proposed model. 
    }
    \label{tab:compare_two}
\resizebox{\columnwidth}{!}{%
\begin{tabular}{@{}l|ccc|ccc@{}}
\toprule
\multirow{2}{*}{Model} & \multicolumn{3}{c|}{Infeasible Steps (\%)} & \multicolumn{3}{c}{Infeasible Trajectories (\%)} \\
 & Acceleration & Curvature & Any & Acceleration & Curvature & Any \\ \midrule
Ground Truth & 0.6 & 9.4 & 10.0 & 11.9  & 25.2  & 28.9  \\
HPTR & 3.7 & 11.9 & 15.6 & 60.5 & 37.2 & 89.1 \\
${\text{HPTR}}_{\,\text{bd}}$ (Ours) & \textbf{0.0} & \textbf{0.0} & \textbf{0.0} & \textbf{0.0} & \textbf{0.0} & \textbf{0.0}\\ \bottomrule
\end{tabular}
}
\vspace{-0.35cm}
\end{table}
\begin{table*}
    \centering
    \resizebox{0.85\textwidth}{!}{
    \begin{tabular}{l|ccccccc}
    \toprule    
    Model & minADE$_6$ & minFDE$_6$ & brier-minADE$_6$ & brier-minFDE$_6$ & minADE$_1$ & minFDE$_1$ & MR$_{2.0}$ \\ \midrule
    HPTR &  7.42 & 20.19 & 8.12 & 20.88 & 10.72 & 30.34 & 1.00 \\
${\text{HPTR}}_{\,\text{kin}}$ & 10.29 & 28.00 & 10.57 & 28.29 & 10.29 & 28.00 & 1.00 \\
${\text{HPTR}}_{\,\text{bd}}$ (Ours) & \textbf{3.11} & \textbf{3.84} & \textbf{3.79} & \textbf{4.52} & \textbf{5.09} & \textbf{8.45} & 1.00 \\ \bottomrule
\end{tabular}}
\caption{
Comparison of plain HPTR, HPTR$_{\,\text{kin}}$, and our model on left U-turn maneuvers using benchmark metrics.}
\label{tab:compare_three}
\end{table*}
\subsection{Generalization to Unseen Road Topologies}
The results from the maneuver analysis motivated a deeper investigation into the generalization and robustness of the proposed model. Specifically, we evaluated off-road rates under adversarial perturbations introduced scene attack~\cite{bahari2022vehicle}. Scene attack applies three types of road topology perturbations—smooth-turn, double-turn, and ripple-road—designed to simulate realistic and physically feasible road topologies, challenging the model's road awareness capabilities. Using 100 scenes from the AV2 validation split, we applied each perturbation type 18 times with varying power levels, following the scene attack methodology.

\begin{figure*}[t]
    \centering
    \begin{subfigure}[t]{0.32\linewidth}
    \centering
    \includegraphics[width=\linewidth]{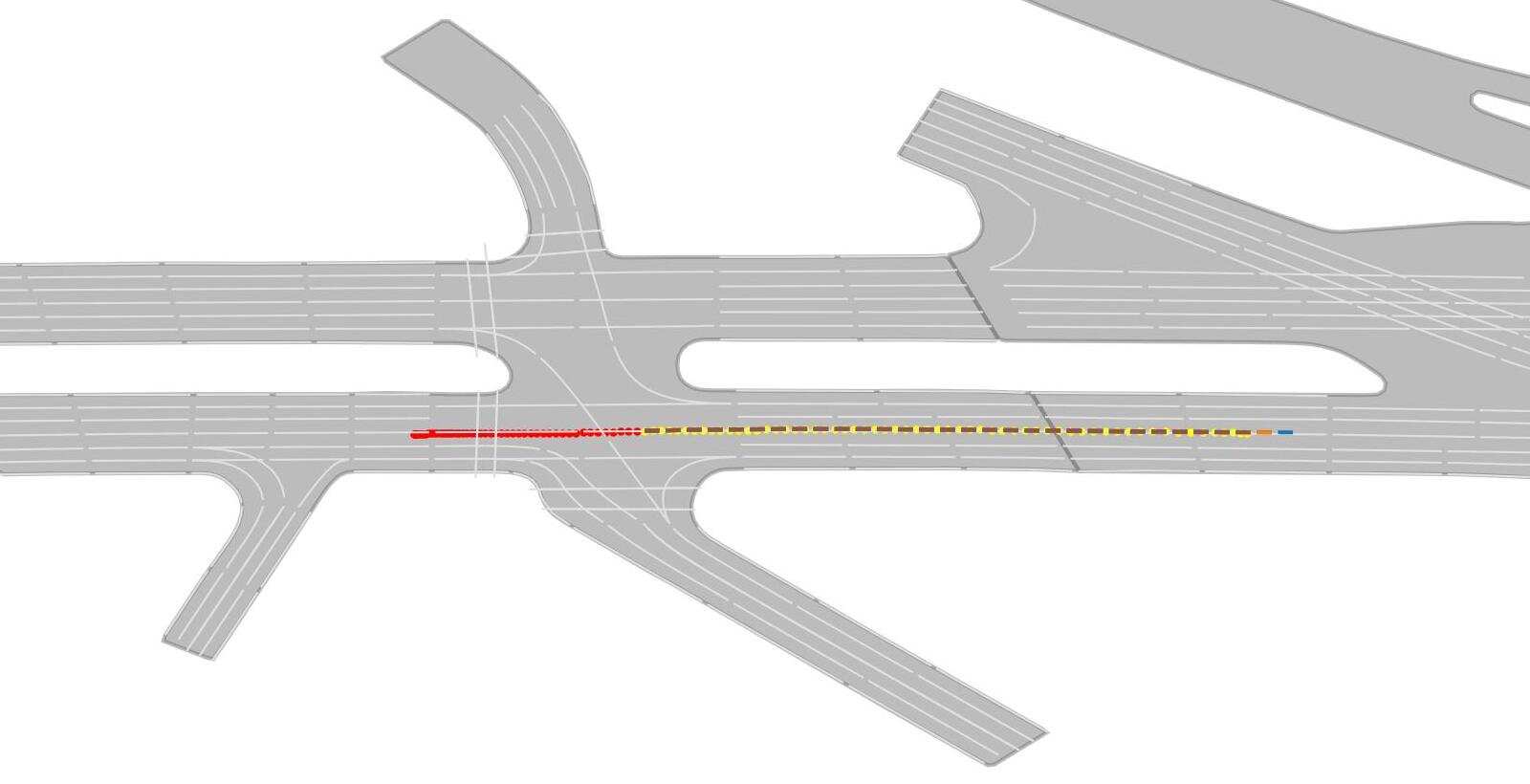}
    \caption{Original Scene}
    \end{subfigure}
    \hfill
    \begin{subfigure}[t]{0.32\linewidth}
    \centering
    \includegraphics[width=\linewidth]{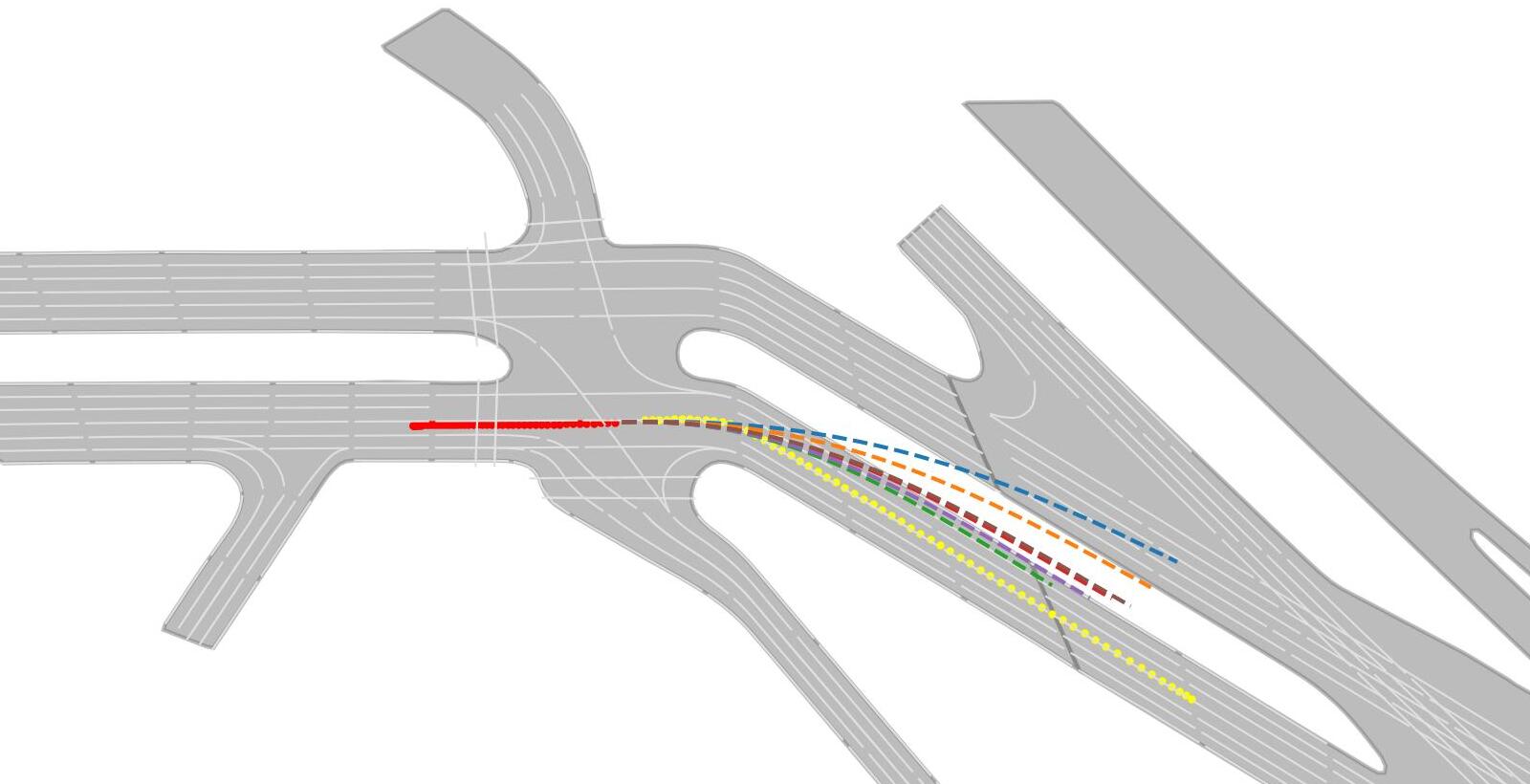}
    \caption{HPTR predictions after a smooth-turn attack}
    \end{subfigure}
    \hfill
    \begin{subfigure}[t]{0.32\linewidth}
    \centering
    \includegraphics[width=\linewidth]{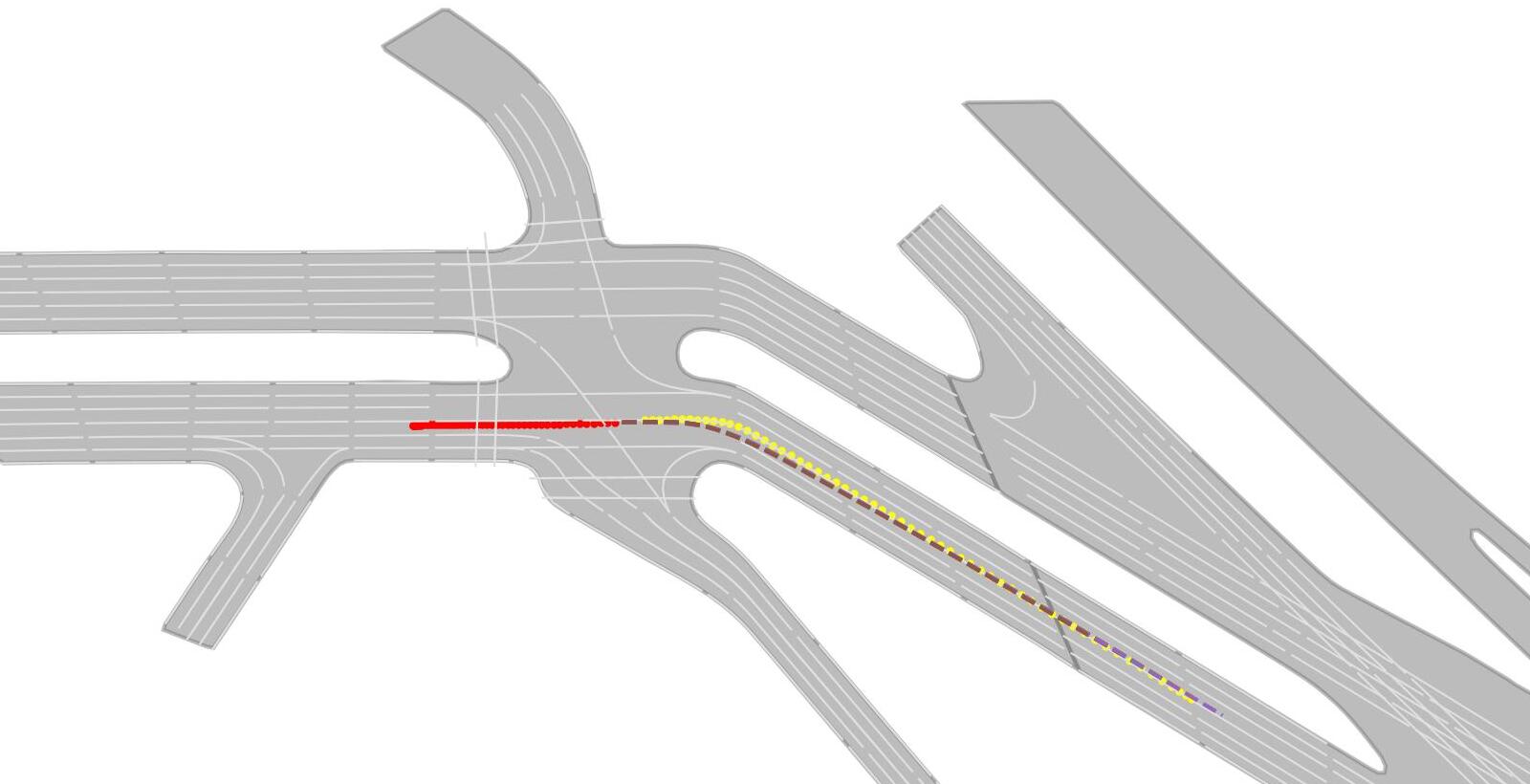}
    \caption{Our predictions after a smooth-turn attack}
    \end{subfigure}
    \caption{HPTR and our model predictions under scene attacks show that HPTR struggles with road topology perturbations, while our approach robustly predicts feasible, on-road trajectories. The focal agent's history is in red, and the ground truth is in yellow.}
    \vspace{-0.2cm}
    \label{fig:smooth_attack}
\end{figure*}

Results reveal significant brittleness in HPTR when exposed to these attacks, with 66\% of trajectories containing at least one off-road point and 26.18\% of all predicted trajectory steps deviating off-roads, as shown in Table~\ref{tab:compare_four}. In contrast, our proposed methodology drastically reduces the off-road rate to just 1.0\%, with only 0.325\% of all predicted steps veering off-road. Further analysis revealed that these rare off-road occurrences, accounting for 0.325\%, result from minor approximation errors introduced during the boundary smoothing step. Fig.~\ref{fig:smooth_attack} illustrates a scene from the AV2 validation split, demonstrating the impact of a smooth-turn attack on model predictions. HPTR fails to generalize to the unseen road topology despite the presence of a feasible ground truth. In contrast, our proposed model with boundary guidance generates reasonable predictions that remain within the road and closely align with the ground truth.
\begin{table*}[]
\renewcommand{\arraystretch}{1.1}
\centering
\caption{Evaluation of the proposed model versus HPTR on scene-attack scenarios, assessing their generalization to unseen road topologies based on off-road percentages (SOR / HOR).}
\resizebox{0.8\linewidth}{!}{
\begin{tabular}{l|c|ccc|c}
\toprule
 Model & \begin{tabular}[c]{@{}c@{}}Original\\ SOR / HOR (\%)\end{tabular} & \begin{tabular}[c]{@{}c@{}}Smooth-turn\\ SOR / HOR (\%)\end{tabular} & \begin{tabular}[c]{@{}c@{}}Double-turn\\ SOR / HOR (\%)\end{tabular} & \begin{tabular}[c]{@{}c@{}}Ripple-road\\ SOR / HOR (\%)\end{tabular} & \begin{tabular}[c]{@{}c@{}}All\\ SOR / HOR (\%)\end{tabular} \\ \midrule
HPTR & 0.35 / 1.5 & 9.93 / 29.5 & 17.37 / 51.5 & 25.04 / 61.0 & 26.18 / 66.0\\
${\text{HPTR}}_{\,\text{bd}}$ (Ours) & \textbf{0.003} / \textbf{0.1667} & \textbf{0.0} / \textbf{0.0} & \textbf{0.28} / \textbf{1.0}  & \textbf{0.05} / \textbf{0.34}  & \textbf{0.325} / \textbf{1.0}  \\ \bottomrule         
\end{tabular}}
\label{tab:compare_four}
\end{table*}
\section{Conclusion}
This paper introduces a novel approach to trajectory prediction that enhances road awareness and physical feasibility in a more flexible manner than previous works. A key contribution is the introduction of a constrained output space through boundary sets, supported by an algorithm that generates permissible driving directions and their boundaries. These sets ensure road alignment while enabling flexibility for complex maneuvers. By representing boundaries as polylines and leveraging attention mechanisms, the network captures spatial and contextual relationships between boundaries, map lanes, and relevant agents. The output heads use a superposition mechanism to combine left and right boundary polylines into a learned path, transformed into a physically feasible, on-road trajectory through a kinematic pure pursuit layer and estimated acceleration profiles. Experimental results demonstrate that our approach significantly reduces off-road and physically infeasible predictions while maintaining competitive accuracy in benchmark metrics. The model generalizes well to unseen road topologies and excels in complex scenarios such as U-turns, highlighting its robustness. Future work may explore extending constrained predictions to pedestrian and cyclist motion, which pose unique challenges due to their non-road-bound movement. We intend to integrate our prediction model with planning systems to further assess its robustness and generalization.

{
    \bibliographystyle{IEEEtran}
    \bibliography{references}
}
\end{document}